\documentclass[conference]{IEEEtran}

\usepackage{booktabs} 
\usepackage{graphicx}
\usepackage[compatibility=false]{caption}
\usepackage{subcaption}
\usepackage{hyperref}
\usepackage{mathtools}
\usepackage[linesnumbered,ruled,vlined]{algorithm2e}
\DeclarePairedDelimiter\ceil{\lceil}{\rceil}

\begin{document}

\title{Performance-Oriented\\ Neural Architecture Search}

\author{\IEEEauthorblockN{Andrew Anderson\IEEEauthorrefmark{1},
Jing Su\IEEEauthorrefmark{2}, Rozenn Dahyot\IEEEauthorrefmark{3} and
David Gregg\IEEEauthorrefmark{4}}
\IEEEauthorblockA{School of Computer Science and Statistics,
Trinity College Dublin\\
Dublin, Ireland\\
Email: \IEEEauthorrefmark{1}andersan@cs.tcd.ie,
\IEEEauthorrefmark{2}jing.su@tcd.ie,
\IEEEauthorrefmark{3}Rozenn.Dahyot@tcd.ie,
\IEEEauthorrefmark{4}dgregg@cs.tcd.ie}}






\maketitle

\begin{abstract} Hardware-Software Co-Design is a highly successful strategy
for improving performance of domain-specific computing systems. We argue for
the application of the same methodology to deep learning; specifically, we
propose to extend neural architecture search with information about the
hardware to ensure that the model designs produced are highly efficient in
addition to the typical criteria around accuracy.

Using the task of keyword spotting in audio on edge computing devices,
we demonstrate that our approach results in neural
architecture that is not only highly accurate, but also efficiently
mapped to the computing platform which will perform the inference.
Using our modified neural architecture search, we demonstrate $0.88\%$ increase 
in TOP-1 accuracy with $1.85\times$ reduction in latency for keyword spotting
in audio on an embedded SoC, and $1.59\times$ on a high-end GPU.
\end{abstract}




\section{Motivation}

For modern edge computing systems, the task of implementing deep learning
applications with high efficiency has become critical. For example, automatic
speech recognition~\cite{DBLP:journals/spm/X12a} (ASR) is an area where
significant industrial effort is being focused to move the problem on-device
and avoid an always-on connection to cloud computing resources.

The typical approach is first to select a deep neural network (DNN)
architecture and train it to acceptable accuracy on some task. Techniques such
as weight pruning~\cite{DBLP:conf/isca/YuLPDDM17} and model
quantization~\cite{DBLP:journals/tcad/WessDJ18} are then applied, so that the
inference has acceptable latency and memory consumption on the target device.
This general workflow is applied in the vast majority of research, for example,
in the recent on-device ASR work at
Google~\cite{DBLP:journals/corr/abs-1811-06621}.

However, certain features of neural architecture, especially in convolutional
neural networks (CNNs) present hard roadblocks for performance. For example,
the size and shapes chosen for convolution kernels determines whether or not
fast and memory-efficient convolution algorithms can be applied at inference
time~\cite{DBLP:conf/cgo/AndersonG18}.

In this paper, we argue for a more flexible and holistic approach to
optimization of deep learning applications. We propose to adapt
techniques from hardware-software co-design to the domain of deep
learning, by performing \textit{neural architecture search} with both
classification accuracy \emph{and} inference performance as primary
objectives.

Extending neural architecture search with information about the hardware on
which the inference will be deployed allows decisions to be made at a high
level concerning the structure of the neural network which have profound
implications for performance. Without this information, we are searching ``in
the dark'' with regard to inference performance, and it is unlikely that the
search will find high efficiency neural architectures, except by chance.

\subsection*{Contribution}

\noindent In this paper, we extend neural architecture search with an inference
performance cost model to allow performance concerns to be
directly integrated in the architecture search.

We propose a neural architecture search based on the idea of iterative refinement of
stepwise-optimal candidates that allows the user to control the depth and 
breadth of the search.

We demonstrate that it is possible to find variations of neural architecture
with equivalent classification accuracy but greatly improved inference
performance, even before weight pruning or model quantization are considered.

\subsection*{Paper Structure}

\noindent We first present some background on neural architecture search, and inference
performance in deep neural networks.

Next, we present our performance-oriented neural architecture search algorithm. 
Using our approach, we explore variations of the initial \emph{seed} 
architecture and collect classification accuracy and inference
performance statistics.

We evaluate our approach on a real-world edge platform by
considering the application of a keyword spotting (KWS) pipeline on an ARM
Cortex-A class processor. For context, we also benchmark the models produced 
by our approach on a typical high-end GPU commonly used for deep-learning 
workloads. Finally, we present some related work and discuss potential 
extensions of our approach.

\section{Background}

Many deep learning models in widespread use today are still designed by
humans~\cite{DBLP:journals/corr/Krizhevsky14,
DBLP:journals/corr/IandolaMAHDK16} or according to simple schemes where
submodules are programmatically \emph{stacked} to a configurable depth
to construct the
network~\cite{DBLP:conf/cvpr/SzegedyLJSRAEVR15,DBLP:journals/corr/WuSH16e,DBLP:journals/corr/SimonyanZ14a}.

Neural architecture search is typically performed in conjunction with
hyperparameter optimization to automatically \emph{discover} model
architectures which can be trained to a high degree of accuracy for a
specific learning task~\cite{DBLP:journals/corr/ZophL16,DBLP:journals/corr/HaDL16, 2018arXiv180805377E}.

However, the search process may discover many model architectures which
can be trained to an approximately equivalent degree of accuracy. When
resources are no object at inference time, any of these models will
suffice. However, when performing inference on edge devices, with
limited memory and compute capacities, it is crucial to optimize the
model to minimize storage and computational requirements.

Variants of neural architecture search have been successfully applied
to many networks in different contexts in prior
work~\cite{DBLP:journals/corr/ZophL16}. However, these investigations
have focused only on improving the \emph{accuracy} of the trained
networks.

Yang et al. propose the use of co-design methods to produce a 
customized hardware design to accelerate a specific neural 
network~\cite{DBLP:conf/fpga/YangHWZ0GBLVWK19}.
Our work differs from the work of Yang et al. primarily in that we
propose a generic method by which the entire space of customized networks 
derived from some initial seed network can automatically be explored, 
rather than designing a custom network by hand.

Cai et al.~\cite{DBLP:journals/corr/abs-1812-00332}
propose a modification to the training process of the neural network
to directly incorporate parameter search. However a key difference with our work
is that Cai et al. require a set of fixed alternatives to be chosen in advance, rather
than our fine-grained bounds for individual architectural parameters.
Nevertheless, the approach of Cai et al. is shown to yield improvements in two image classification tasks.

\emph{Pruning} weights with small values induces sparsity in
the model~\cite{DBLP:conf/isca/HanLMPPHD16} and is a popular technique for
decreasing operation count. However, the degree to which a model can be pruned without failing to
meet classification accuracy targets is unpredictable and largely
model-dependent. Furthermore, it is not always clear how to select an effective
pruning approach for any given model beyond trial and error.

\emph{Quantization} is another major approach to reducing model size
and improving inference performance. By converting weights to a smaller
numeric format, the model size is reduced. Although quantization does
not change the number of operations, computation on smaller types
is often faster, especially on FPGA~\cite{DBLP:journals/trets/BlottPFGOULV18},
or custom accelerators~\cite{DBLP:conf/isca/HanLMPPHD16},
where the arithmetic implementation in hardware can be customized 
to match the model quantization scheme.

On general purpose processors and GPUs, quantization is restricted to 
the use of the numeric types already present in hardware. Nonetheless,
quantization often results in a net gain in inference performance~\cite{DBLP:conf/cf/PradoDBP18}.

\subsection{DNN Convolution}

Convolution is the most computationally heavy primitive in the majority
of popular deep neural networks. As such, it presents a natural target
for optimization.

\begin{figure}
\centering
\includegraphics[scale=0.15]{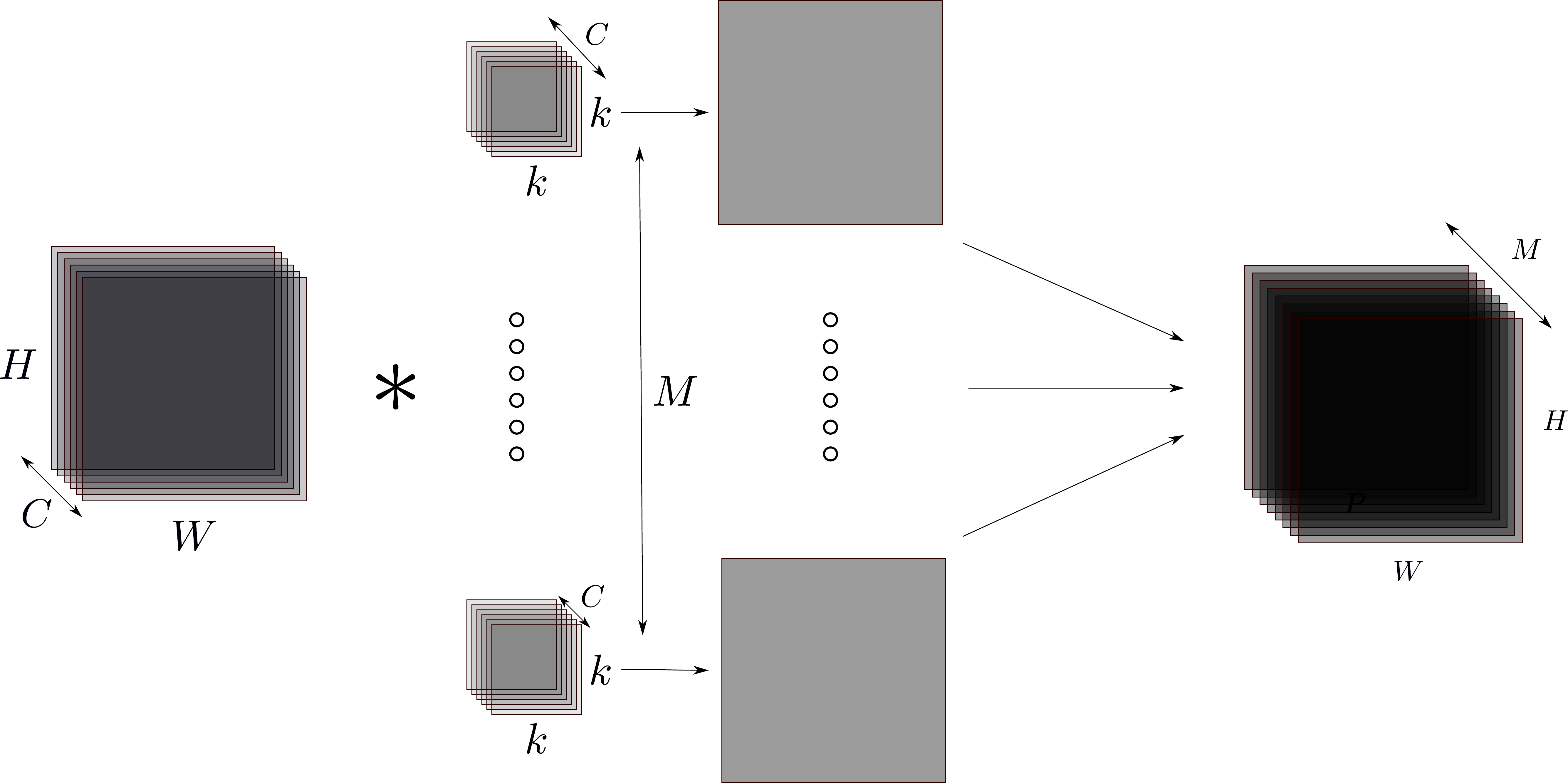}

\caption{\textbf{DNN Convolution}: $C$ input
feature maps, each of size $H \times W$, are convolved with $M$ multichannel
filters, each with $C$ channels and a $k \times k$
kernel, resulting in $M$ output feature maps.}

\label{fig:mcmk}
\end{figure}

\noindent Figure~\ref{fig:mcmk} shows the data flow in the DNN
convolution operation. The $H$ and $W$ parameters are determined by the
size of the input feature maps, but the $k$, $C$, and $M$ parameters
characterize the convolution being applied. The \emph{filter size} of
the convolution is $k \times k$ points, and the number of feature maps,
$C$, determines how many of these $k \times k$ filters are trained.
Since convolutions are connected input-to-output in the neural network,
the number of feature maps on the output, $M$, matches the number of
feature maps $C$ on the input of the next convolution in the chain.

Since inference performance is dominated by the convolutional layers,
the size and number of convolutional filters trained in each
convolutional layer are natural candidates for search. With very few or
very small filters, the network encodes much less information than with
more numerous or larger filters. However, a reduction in filter
quantity or filter size reduces both the operation count, and the model
size, resulting in more efficient inference.

Striking a balance between these opposing objectives is the goal of
this paper. We propose to incorporate inference performance concerns
directly in the neural architecture search, by extending neural
architecture search with a new objective function incorporating
inference performance.

\subsubsection*{Inference Performance}

\noindent The inference performance of convolutional DNNs is dominated by the time spent
executing the convolutional layers. $k_h k_w$ multiplications and additions are
used to compute the output of a single filter. All $C$ filters are then applied
at $H \times W$ input points, where $H$ and $W$ are subdivided by their
respective strides, $S_h, S_w$, if present. $C$ additions per output point are
then performed to compute one output feature map. Since there are $M$ output
feature maps, the total operation count is
$M(C\ceil{\frac{H}{S_h}}\ceil{\frac{W}{S_w}}
(2 k_h k_w) + C\ceil{\frac{H}{S_h}}\ceil{\frac{W}{S_w}})$ or more succinctly,

\begin{equation}
\label{eqn:flops}
FP_{OPS} = MC\ceil*{\frac{H}{S_h}}\ceil*{\frac{W}{S_w}} (2 k_h k_w + 1)
\end{equation}

\noindent Similiarly, the number of weight parameters for a convolutional layer
can be expressed as

\begin{equation}
\label{eqn:weights}
FP_{PARAMS} = MC(k_h k_w)
\end{equation}

%

\noindent There are many special-case algorithms for computing a direct convolution,
which all have the same asymptotic complexity, but exploit spatial locality and
other features of the convolution to make efficient use of different primitive
operations, such as matrix multiplication~\cite{DBLP:conf/asap/VasudevanAG17}. The choice of filter size and shape determines which of these algorithms can actually be used.

\section{Our Approach}

Prior work~\cite{DBLP:journals/corr/abs-1708-05344,
DBLP:journals/corr/ZophL16} on neural architecture search ranges in
scope from the design of whole-network connective structure to the
value ranges of specific hyperparameters. In order to demonstrate our
performance-oriented search, we focus on the two \emph{kinds} of
parameters which most immediately influence operation count in the
network; filter size, $k$, and feature map count, $M$. We consider the
filter size $k$ to consist of two distinct sub-parameters, $k_h$ and
$k_w$, to allow for non-square filters, since these are observed in
several expert-designed networks for keyword spotting.

Taking the expert-designed network from the work of De Prado et
al.~\cite{DBLP:conf/cf/PradoDBP18} as the initial \emph{seed}
architecture, we explore variant networks with the same connective structure,
but different $k_h$, $k_w$, and $M$ parameters on all layers.

In principle, this search can be extended to incorporate many other parameters, provided that
their impact on the operation count in the network can be specified.

\begin{figure*}
\centering
\includegraphics[width=\textwidth]{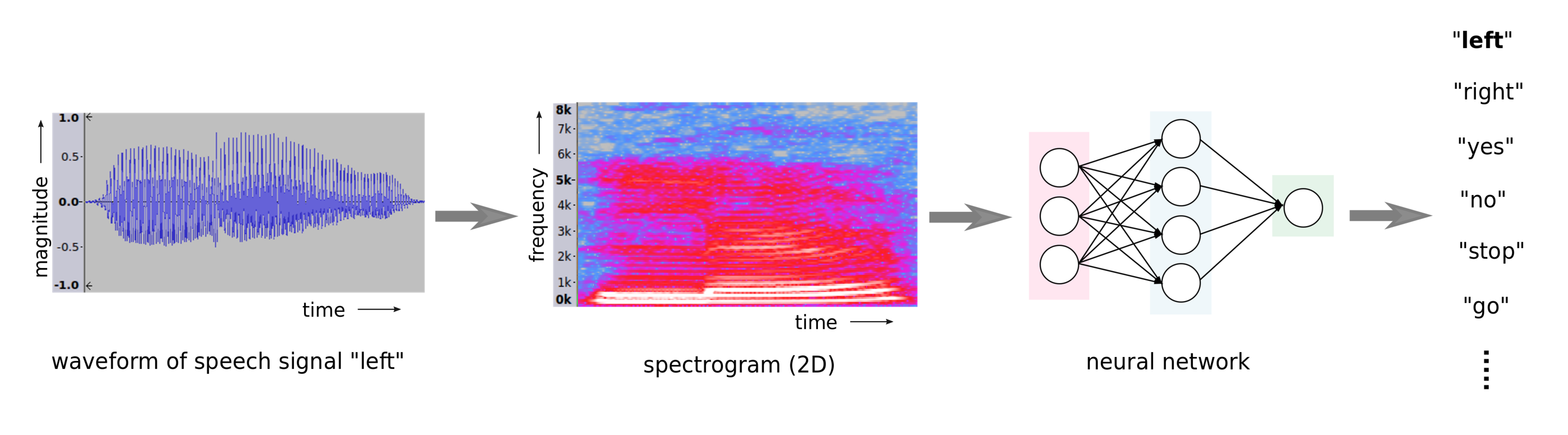}
\caption{\textbf{Keyword Spotting workflow.} A one-second mono-channel speech signal is used as input. An MFCC spectrogram is computed and used as the input to a neural network. At inference, an unknown signal is processed through this workflow to produce a label prediction.}
\label{fig:kws}
\end{figure*}

\begin{figure}
\centering
\includegraphics[width=0.95\linewidth]{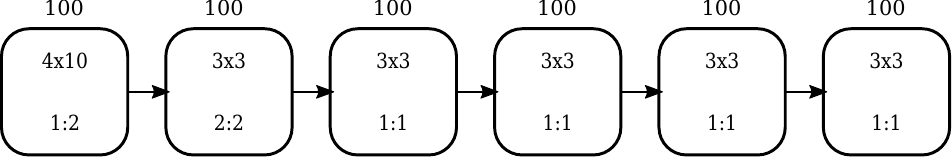}

\caption{\textbf{Seed CNN Architecture}. Each unit performs a
convolution, batch normalization, scale, and ReLU activation. The first and
second unit perform strided convolutions with the indicated vertical and horizontal subsampling factors, but the remaining convolutions are
stride 1. All units have 100 filters of size $3\times3$ except the first
unit, where the filter size is $4\times10$.}

\label{fig:original-cnn}
\end{figure}

\subsection{Search}

\noindent We begin by constructing a search space by fixing a range for each of the
parameters $k_h$, $k_w$ and $M$, for each convolution in the network. Each
point in this search space is a unique configuration of the neural network.
There are two key metrics of network quality which we consider: accuracy and
cost. To model how well any given configuration of the network classifies, we
measure the TOP-1 accuracy across some fixed number of inferences. Using the
formula in Equation~\ref{eqn:flops}, we also compute the cost of the
configuration as the sum of the operation count in all convolutional layers.

In principle, any measure of accuracy and of cost can be used. For
example, to relax the accuracy requirement, the average top-5 accuracy
could be considered instead of TOP-1.

Since $k_h$, $k_w$, and $M$ are arbitrary (positive) integers, the
configuration space of the network is extremely large. Recall that these
parameters can vary independently for each convolutional layer in the network.
In practice, for any one layer, $k_h$ and $k_w$ are bounded above by the height
$H$ and width $W$ of the input feature maps, since setting $k_h = H$ and $k_w =
W$ means that the convolution produces a single scalar output. However the
number of filters $M$ is unbounded in practice, although we would expect
diminishing returns in terms of accuracy as the filter count grows very large.

Even by fixing $k_h$, $k_w$, and $M$ to be bounded above by their settings in
the \emph{seed} network, many configurations remain, the evaluation of which
requires some number of training iterations.

Prior work on Neural Architecture Search has employed numerous approaches to
deal with the size and complexity of the search spaces involved. All of the
approaches share a common design element, in that they attempt to learn the structure
of the search space so that the majority of the search time is spent evaluating
candidate configurations which are likely to be of high quality.

Two of the most popular approaches are to use \emph{reinforcement learning} or
\emph{hypernetworks} to learn the structure of the search space. In the ``SMASH'' approach of
Brock et al.~\cite{DBLP:journals/corr/abs-1708-05344}, an estimator is
trained using reinforcement learning which proposes new points in the space
that are likely to be high-quality configurations. In the
approach of Le et al.~\cite{DBLP:journals/corr/ZophL16}, a recurrent neural
network (RNN) is trained, which learns to propose strings specifying the
configuration of the network. This approach works by feeding the accuracy of
the child network (the CNN matching the proposed configuration) back into the
parent RNN to modify the loss.

In principle, any method may be chosen to
explore the configuration space of the network. Regardless of the choice of
method, the ideal result is a set of configurations of the network which are
high-quality, in the sense that they have high accuracy and low cost. However,
according to the needs of the programmer, some of these high-quality
configurations may be better than others. For example, if the aim of the search
is strictly to find the least-cost network which meets some accuracy target, a
different candidate may be best for a search aimed at finding the most
accurate network that does not exceed a certain cost.

To evaluate a candidate model while exploring 
the search space, some budget of training iterations is required. 
Depending on the available computing resources for training,
this may be larger or smaller. With a smaller training iteration budget, more points in the search space can be explored in a
fixed amount of time than with 
a larger budget. However, with a larger budget, the candidates will be evaluated in more depth, meaning discriminating between them will be easier.  
The stopping criterion for one search phase may be a wall-clock
time elapsed, or a certain number of models explored. In either case, after 
executing a search phase, we obtain a set of \emph{candidate} models trained
for some number of iterations.

\subsection{Refinement}

\noindent The second phase of our approach is the \emph{refinement} of the set of
candidate models. We construct the \emph{Pareto frontier} of the model
set produced from the search stage with respect to the two criteria of accuracy
and cost. 

If all, or very many, candidates on the Pareto frontier share a common
substructure (i.e. common settings for any $k_h$, $k_w$, or $M$ parameter),
we fix that substructure, and repeat the search phase. Since
we have reduced the search space by fixing some substructure, each of these
iterative refinements becomes faster to evaluate. This means that we can
explore more of the search space around refined configurations (with a fixed training iteration budget), or evaluate candidates more precisely (with an increased training iteration budget). In this way, the search
process works at a progressively finer granularity the more refinement steps
are taken.

A Pareto improvement is a change that makes at least one criterion better without making any other criterion worse, given a certain initial configuration. A configuration is \emph{Pareto optimal} when no further Pareto improvements can be made. The Pareto frontier is the set of all Pareto optimal configurations, i.e. configurations of the network where
no other configuration can achieve at least as much accuracy with lower cost, nor 
strictly more accuracy with equivalent cost.

In practice, we found that fixing the \emph{most commonly observed parameter setting}
in the Pareto-optimal models worked very well in the refinement phase.
It is important to note that when we freeze a parameter setting, although we do not continue to evaluate candidates 
which do not share the setting for the fixed architectural parameter, we do not \emph{discard} Pareto optimal candidates which have already been found. Models on the Pareto frontier which do not share the chosen setting
for a frozen parameter might continue to be on the frontier even after the other candidates,
and their variants, have been trained for further iterations.
Algorithm \ref{algo:nas} summarizes the workflow of our neural architecture search strategy. Steps 4 and 5 may be repeated until all architectural parameters are frozen.

\begin{algorithm}
\SetAlgoLined
 Select the seed architecture \\
 Initial search over network/optimizer parameters \\
 Select Pareto-optimal architectures \\
 Refined search over network parameters \\
  Select Pareto-optimal architectures and freeze parameters \\
 Fine-tune Pareto-optimal models \\
\caption{Neural Architecture Search}
\label{algo:nas}
\end{algorithm}

\section{Experimentation}

In this paper, we use keyword spotting (KWS) as an example to study neural architecture search. A KWS system takes an input audio signal and predicts 
its most likely text label. This system can be trained to work through a typical 
supervised learning approach together with labelled speech samples. Instead of 
classifying audio signals directly, a feature extraction step is applied to generate 
a spectrogram from the audio. Computing the Mel-frequency Cepstral Coefficients (MFCCs)~\cite{Davis:1980tg} 
is a typical preprocessing step for ASR applications. Figure~\ref{fig:kws} 
shows an end-to-end KWS workflow in which 2D MFCCs are used to train a neural 
network classifier. For all models we generate MFCC features in the same way. 
40 frequency bands are applied to 16kHz audio samples with 128ms frame length and 
32ms stride. We use the Google speech command data set ~\cite{Warden2018} 
which has samples of 1 second in length, so the MFCC tensor has 40x32 features.

The speech commands dataset contains 65,000 audio samples of 30 keywords. We 
selected 10 keywords (\emph{yes}, \emph{no}, \emph{up}, \emph{down}, \emph{left}, 
\emph{right}, \emph{on}, \emph{off}, \emph{stop}, \emph{go}) for 
our experiment and used the default training, validation and test split, 80:10:10 respectively.

We implemented our approach using the Microsoft NNI (Neural Network
Intelligence) toolkit~\cite{MSNNI}, for constructing automated machine learning
(AutoML) experiments. NNI is designed for building automated searches for the
best neural architecture and/or hyper-parameter sets.

For the two objectives we chose the TOP-1 accuracy across 100 inferences as the
estimator of network accuracy, and the sum of operation counts
(Equation~\ref{eqn:flops}) as the estimator of network cost. Rather than a
reinforcement learning or RNN based search, we chose to use the Tree-structured
Parzen Estimator~\cite{DBLP:conf/nips/BergstraBBK11} (TPE), since each trial
involves training a neural network to convergence, so the trials are high-cost.
This estimator is already implemented in the NNI toolkit. 


\subsection*{Seed Architecture}

A 6 layer CNN model (Figure~\ref{fig:original-cnn}) previously achieved 94.23\% test accuracy 
with Google speech commands dataset~\cite{Warden2018}, therefore 
it is a good starting point of neural architecture search. 
We refer to the Caffe-trained model of~\cite{DBLP:conf/cf/PradoDBP18}
 as the \textit{original model} (i.e. with the parameters as found in \cite{DBLP:conf/cf/PradoDBP18} that yields 94.23\% TOP-1 accuracy), and its neural architecture as the \emph{seed} architecture. 


Starting with the seed architecture, we configured NNI to perform a search for the 
number of feature maps $M$, kernel height $k_h$ and kernel width $k_w$ for each 
convolutional layer.


\subsection*{Search in Experiments}

\noindent Microsoft's NNI toolkit offers many options to tune network hyperparameters as well as 
solver/optimizer parameters, including training iterations, batch size, 
learning rate and decay strategy. However, the search space 
expands exponentially if we try to search for values for all parameters independently. 
In order to finish our experiment within a reasonable time budget we used a two step approach. 
Firstly, we explored solver parameters used to train the original model. The original model is trained in
40,000 iterations with the ADAM 
optimizer~\cite{kingma:adam} with a batch size of 100. The base learning rate is $5\times10^{-4}$ 
and it drops 70\% every 10,000 iterations. 

Performing a hyperparameter search with NNI, we found that using a learning rate of $1\times10^{-3}$ and batch size of 25 with 8,000 training iterations had high correlation with top accuracy scores. The ADAM optimizer was still preferred. We fixed this set of new solver parameters before beginning our search.

Our search space bounds for the three parameter types were configured as follows.
For the per-layer bound on $M$, we used the setting in the seed architecture
$M=100$ as an upper bound, with a lower bound of $1$. For the settings of $k_h$ and $k_w$, we used the
seed architecture settings as upper bounds for the first convolutional layer
($k_h = 4, k_w = 10$), and for the remaining layers we increased the bounds
slightly from the original setting of ($k_h = 3, k_w = 3$) to ($k_h = 5, k_w =
5$). All $k_h$ and $k_w$ lower bounds were also set to $1$.

This choice of bounds means we are searching for networks where each layer has
at most as many filters, $M$, as in the original network, and the first layer
filter size is at most as large as in the original network. The filter sizes of
subsequent layers may be larger than in the original network, up to a bound of
$66\%$ growth in each spatial dimension.

\subsection*{Refinement in Experiments}

\begin{figure*}
\centering
\includegraphics[width=0.7\textwidth]{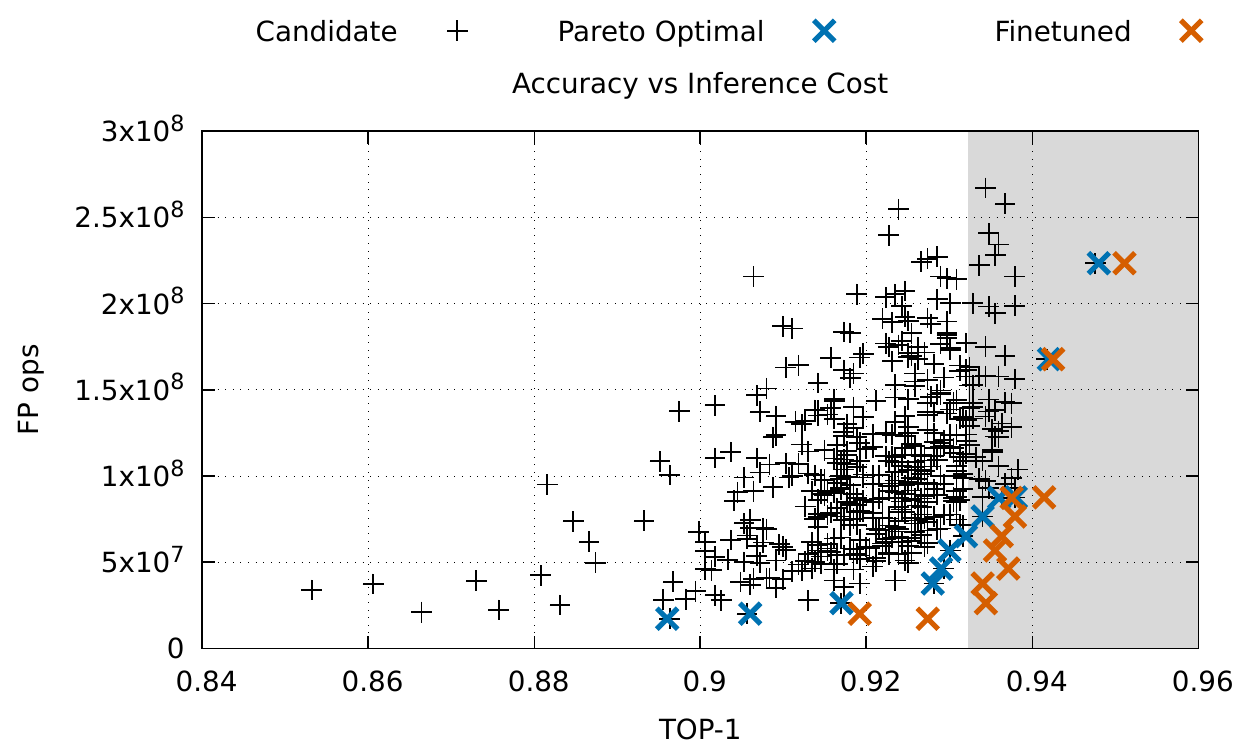}

\caption{\textbf{Experimental Results from Model Search}.\\ 
The figure shows the operation count of the candidate models versus their TOP-1 test 
accuracy. Models within at most -1\%
point deviation from the test accuracy of the \emph{seed} model occupy the
shaded gray area in the graph. Pareto-optimal models are indicated with crosses $\times$. All models were trained for 8,000 iterations during search. The final set of Pareto-optimal models were trained for a total of 40,000 iterations (equivalent to the \emph{seed} model). These are the ``Finetuned'' data points. }

\label{fig:experiment}
\end{figure*}

\noindent We ran the search stage of the experiment with our initial configuration until
300 candidate networks were produced. Looking at the Pareto frontier in this
experiment, we noticed that the vast majority of high-quality candidates used a
filter size of ($k_h = 3, k_w = 3$) for the first convolutional layer. There
was no other significant common substructure in the candidates. Fixing
this filter size for the first layer, we repeated the search step with the
refined \emph{seed} network until a further 500 candidates were produced.
We computed the Pareto frontier for this refined set of candidates, and 
fine-tuned the 12 Pareto-optimal models until each had hit the limit of 40,000 training
iterations (i.e. the number of iterations used to train the \emph{seed} model).
Figure~\ref{fig:experiment} shows the results of the experiment.

Since the non-square kernel ($k_h = 4, k_w = 10$) of the first convolutional layer 
yielded high accuracy in CNN, DSCNN and CRNN scenarios~\cite{DBLP:journals/corr/abs-1711-07128},
it is believed to be a good design for KWS. However, our search found that a 
($k_h = 3, k_w = 3$) kernel in the first layer is a better choice.

After investigating the data, we found that the MFCC features were generated from 40ms speech frames~\cite{DBLP:journals/corr/abs-1711-07128}, but we generated MFCC features 
from 128ms speech frames. Consequently the MFCC feature map in our experiment 
covers more temporal information ($W$ dimension of MFCC) than the one in previous work. This setting 
enables good performance of smaller square kernels and it explains why the $k_h$ and $k_w$ values
we observe are preferred by the search. It also demonstrates the power of neural architecture
search to adapt the model to changes in the conditioning of the dataset without intervention
on the part of the end-user.

\subsection*{Observations}

\noindent The most immediate observation from the experimental data is that, once the
TOP-1 accuracy exceeds 90\%, the vast majority of candidates which achieve any
given accuracy target are much more expensive than they need to be. For many of
the gradations in TOP-1 accuracy, the difference between the least cost and
greatest cost configuration is close to, or exceeds, an order of magnitude.

Using traditional approaches to Neural Architecture Search, which are oblivious to
inference performance, we have no way to ensure that the search process will
choose a candidate with reasonable cost, and it is clear that the likelihood of
making a very expensive choice is high.

\begin{table}
\centering
\caption{Interesting Model Configurations from Figure~\ref{fig:experiment}}
\begin{footnotesize}
\begin{tabular}{@{}cccc|c@{}}
\toprule
TOP-1 & M$FP_{OPS}$   & $\delta$ TOP-1 & $FP_{OPS}$   & Note \\ \midrule
0.9423 & 581.12 & 0.0\% & 1$\times$     & seed model             \\ \midrule
0.8960 & 17.22 & -4.63\% & -33.75$\times$  & best $\delta$ $FP_{OPS}$             \\
0.9410 & 87.61 & -0.09\% & -6.63$\times$   & fastest $\delta$ TOP-1 $\approx$ 0       \\
0.9425 & 167.68 & 0.02\% & -3.47$\times$   & fastest $\delta$ TOP-1 $> 0$       \\
0.9511 & 223.44 & 0.88\% & -2.60$\times$  & best $\delta$ TOP-1               \\ \bottomrule

\end{tabular}
\end{footnotesize}
\label{tab:pareto}
\end{table}

\noindent Table~\ref{tab:pareto} summarizes the most interesting models
we found in our experiment. 
The first row shows the configuration which had the largest reduction in operation count versus the
\emph{seed} architecture. The search finds a configuration which uses
$33.75\times$ fewer operations for inference than the \emph{seed} architecture, at the
cost of a $4.63\%$ point reduction in TOP-1 accuracy.

The least-cost configuration found which had approximately the same accuracy as the \emph{seed}
architecture exhibited a reduction in operation count of $6.63\times$, while the least-cost architecture that
was strictly \emph{more} accurate has a reduction of $3.47\times$. Finally, the
most accurate configuration found improves TOP-1 accuracy by $0.88\%$ points, to
$95.11\%$ TOP-1, while reducing operation count by $2.6\times$ over the
\emph{seed} architecture.

\subsection*{Real-World Benchmarking}

\noindent Using the full set of Pareto-optimal CNN models from Figure~\ref{fig:experiment},
we performed two real-world
benchmarking experiments, one on an embedded SoC, the Samsung Exynos 5
Octa (Exynos 5422), and one on a high-end GPU (the NVIDIA GTX 1080Ti).
We used the ODroid XU3
system, which has 2GB of RAM. Our experiments used OpenBLAS 0.3.5.
For GPU, experiments, we used the CUDA toolkit version 10.0 and cuDNN version
7.5.

All models in the Pareto frontier
were benchmarked using Caffe. 100 inferences were performed in all cases.
The runtime of the \emph{seed} model was also benchmarked. Labels on bars
are the TOP-1 score of the model for which inference time is being benchmarked.
All models (including \emph{seed} model) were trained for a total of 40,000 iterations.

\begin{figure}
\centering
\includegraphics[width=\linewidth]{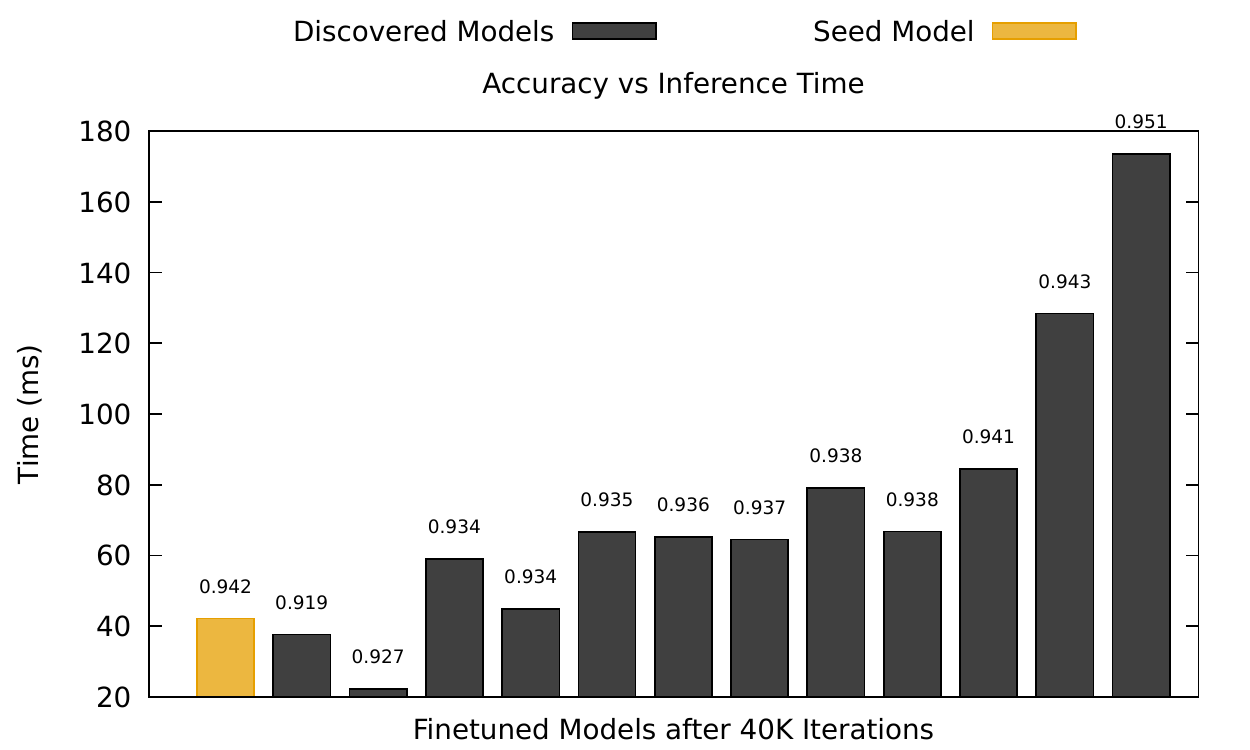}
\caption{\textbf{Exynos 5422 Experiment\\ Single-Inference time (lower is better)} \\
Using our approach, we are able to find models
which are faster (lower latency) than the original \emph{seed} model,
as well as models with better TOP-1 accuracy.
The best speedup observed in practice on the Exynos 5422 is $1.85\times$ versus the \emph{seed} model. Bar labels show TOP-1 accuracy.}
\label{fig:arm-experiment}
\end{figure}

\begin{figure}
\centering
\includegraphics[width=\linewidth]{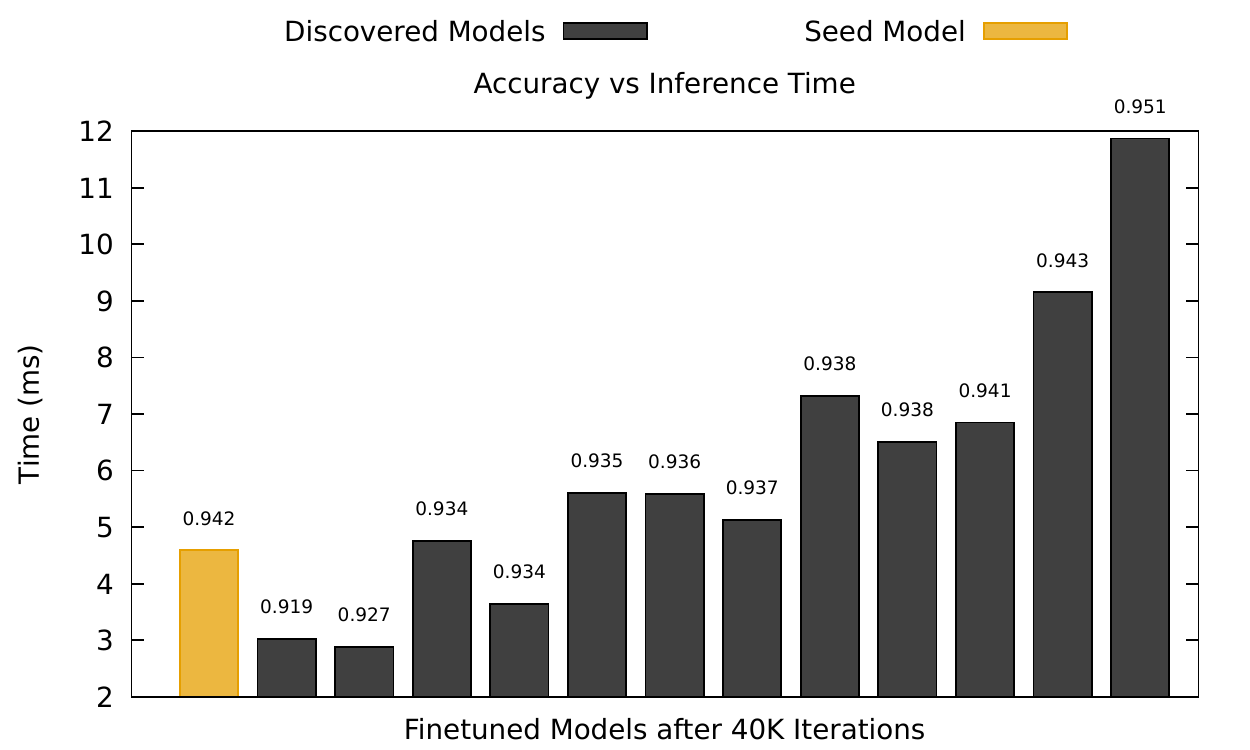}
\caption{\textbf{GTX 1080Ti Experiment\\ Single-Inference time (lower is better)} \\
On the GTX 1080Ti, the spread of inference times is greatly reduced versus the Exynos 5422, due to the high degree of arithmetic throughput available on the GPU. The best speedup observed here is $1.59\times$ versus the \emph{seed} model.  Bar labels show TOP-1 accuracy.}
\label{fig:nvidia-experiment}
\end{figure}

Figure~\ref{fig:arm-experiment} shows the single-inference latency of 
the Pareto-optimal models using Caffe~\cite{jia2014caffe} on the 
Exynos 5422 system, and Figure~\ref{fig:nvidia-experiment} shows the same data on the GTX 1080Ti, using the Caffe cuDNN backend. 

The mapping from $FP_{ops}$ to inference latency is not one-to-one, since Caffe uses GEMM to implement the arithmetic, and spatial locality and other implementation concerns affect the execution time. Nevertheless, we see that the general trend observed in Figure~\ref{fig:experiment} is still observed in practice.

\begin{table*}
\centering
\caption{Pareto Optimal CNN architectures from Figure~\ref{fig:experiment}}
\begin{tabular}{ rcccccccc }

Model & conv1 & conv2 & conv3 & conv4 & conv5 & conv6 & TOP-1 & M$FP_{ops}$    \\
\toprule
seed & 4x10, 100 & 3x3, 100 & 3x3, 100 & 3x3, 100 & 3x3, 100 & 3x3, 100 & 94.2\% & 581.1 \\
\midrule
kws1 & 3x3, 40 & 3x3, 30 & 1x1, 30 & 5x5, 50 & 5x5, 50 & 5x5, 50 & 95.1\% & 223.4 \\
kws2 & 5x5, 40 & 3x3, 50 & 1x1, 30 & 5x5, 40 & 3x3, 50 & 5x5, 50 & 94.3\% & 167.7 \\
kws3 & 5x5, 50 & 1x1, 30 & 5x5, 40 & 3x3, 20 & 5x5, 30 & 3x3, 50 & 94.1\% & 87.6  \\
kws4 & 5x5, 50 & 3x3, 40 & 5x5, 20 & 1x1, 20 & 5x5, 30 & 3x3, 50 & 93.8\% & 87.2  \\
kws5 & 5x5, 20 & 1x1, 40 & 5x5, 30 & 3x3, 20 & 5x5, 30 & 3x3, 30 & 93.8\% & 76.5  \\
kws6 & 5x5, 20 & 3x3, 40 & 3x3, 40 & 3x3, 20 & 3x3, 40 & 3x3, 40 & 93.6\% & 65.2  \\
kws7 & 3x3, 50 & 1x1, 30 & 3x3, 20 & 5x5, 20 & 3x3, 50 & 3x3, 40 & 93.6\% & 56.8 \\
kws8 & 5x5, 50 & 1x1, 50 & 3x3, 20 & 3x3, 40 & 3x3, 30 & 3x3, 20 & 93.7\% & 46.3 \\
kws9 & 5x5, 50 & 1x1, 20 & 1x1, 50 & 3x3, 20 & 5x5, 20 & 3x3, 40 & 93.4\% & 37.7 \\
kws10 & 3x3, 40 & 1x1, 20 & 1x1, 20 & 3x3, 20 & 5x5, 20 & 3x3, 30 & 93.4\% & 26.3 \\
kws11 & 5x5, 30 & 1x1, 20 & 1x1, 20 & 1x1, 20 & 3x3, 20 & 5x5, 20 & 91.9\% & 20.2 \\
kws12 & 5x5, 50 & 1x1, 40 & 1x1, 50 & 1x1, 20 & 3x3, 20 & 3x3, 20 & 92.7\% & 17.2 \\
\bottomrule
\end{tabular}
\label{tab:pareto_best}
\end{table*}

\subsection*{Discovered Models}

Table~\ref{tab:pareto_best} shows the network architectures on the Pareto frontier in Figure~\ref{fig:experiment}. The configuration of convolutional layers is written $\{k_h\times k_w, M\}$.
Final TOP-1 scores after fine-tuning are shown, along with operation count. Additionally, the first row
shows the parameters of the \emph{seed} architecture.

One of the clearest trends in the data is that the choice of 100 filters 
per convolutional unit in the expert-designed seed architecture is excessive. None of the architectures found during search have any unit
with more than 50 filters.

Another trend is that the choice of a $3\times3$ filter for all but the first
unit in the network is suboptimal. In fact, only one of the final
set of Pareto optimal models uses this arrangement (model 6). Looking at 
the most accurate model found (model 1), we see that a smaller 
filter size of $1\times1$ is found for the third unit, while larger $5\times5$ filters
are selected for the last three units. While the choice of 100 filters 
per unit is clearly excessive, it appears that the choice of a $3\times3$ filter size is sometimes too large, and sometimes too small.


Additionally, the choice of a $4\times10$ filter in the first layer, which evenly divides the dimensions of the input spectrogram seems
to be totally unnecessary in our experiments -- not a single Pareto-optimal arrangement of the network uses this filter shape. This expert designed sub-structure could be useful with certain MFCC generation settings but it is not always true. The NAS approach helps us avoid sub-optimal architectures.

The smallest model found, model 12, uses a $5\times5$ filter for the first convolutional unit, $1\times1$ filters for the next three convolutional units, and $3\times3$ filters for the last two units. Despite the enormous reduction in operation count versus the seed model, this arrangement is only $1.5\%$ points less accurate with the same training budget.

\section{Conclusion and Future Work}

It is clear that the design of deep neural networks has become a task
for which pen and paper are no longer suited. We have presented a
computer-aided approach for the design of deep networks, which extends
typical neural architecture search, with a new objective modelling
inference performance. From an initial seed architecture designed by
hand, we are able to automatically discover variants which are as much
as $33.75\times$ more efficient, or as much as $0.88\%$ points more
accurate, as well as the set of Pareto-optimal models spanning these
two extremes. As our evaluation shows, the benefits translate into 
practical performance gains, both on resource constrained embedded devices 
($1.85\times$ speedup) right up to high-performance GPU
systems ($1.59\times$ speedup).

\subsection*{Future Work}
\noindent We have not evaluated the use of pruning or quantization in conjunction with our 
approach. Pruning and quantization apply to a fixed network architecture, and so are orthogonal approaches to reducing model size and operation count. 

As we have demonstrated, using a more appropriate neural architecture 
can result in gains of up to $33.75\times$ reduction in operation count. However, the resulting
model can potentially still benefit from quantization and pruning, and the benefit is cumulative, since the pruning and quantization would be applied to the models which are discovered by search. The evaluation of pruning and quantization in conjunction with model search is a
promising avenue for future work.

\subsection*{Acknowledgments}
\noindent 
This work was  partly supported by Science Foundation Ireland with grants 12/IA/1381,  13/RC/2094  (the Irish Software Research Centre www.lero.ie) and 13/RC/2106 (the Adapt Research Centre www.adaptcentre.ie).
 The project has received funding from the European Union's Horizon 2020 research and innovation programme under grant agreement No 732204
(Bonseyes). This work is supported by the Swiss State Secretariat for
Education, Research and Innovation (SERI) under contract number 16.0159. 

\noindent 
The opinions expressed and arguments employed herein do not 
necessarily reflect the official views of these funding bodies. 

\bibliographystyle{plain}
\bibliography{paper}

\end{document}